\documentclass[letterpaper, 10 pt, conference]{ieeeconf}  

\IEEEoverridecommandlockouts                              

\usepackage{graphicx}
\usepackage[font=small]{caption}
\usepackage{multirow}
\usepackage{tabularx}
\usepackage{cite} 
\usepackage{amsmath}
\usepackage{comment}

\usepackage{siunitx}
\usepackage{multirow}
\usepackage{array}
\usepackage{colortbl} 

\usepackage{siunitx}
\title{\LARGE \bf
Automated Pest Counting in Water Traps through Active Robotic Stirring for Occlusion Handling
}

\author{Xumin Gao$^{*}$\thanks{Corresponding author}, Mark Stevens$^{}$, Grzegorz Cielniak$^{}$%
}
\begin{document}

\maketitle
\thispagestyle{empty}
\pagestyle{empty}


\begin{abstract}

Existing image-based pest counting methods rely on single static images and often produce inaccurate results under occlusion. To address this issue, this paper proposes an automated pest counting method in water traps through active robotic stirring. First, an automated robotic arm-based stirring system is developed to redistribute pests and reveal occluded individuals for counting. Then, the effects of different stirring patterns on pest counting performance are investigated. Six stirring patterns are designed and evaluated across different pest density scenarios to identify the optimal one. Finally, a heuristic counting confidence-driven closed-loop control system is proposed for adaptive-speed robotic stirring, adjusting the stirring speed based on the average change rate of counting confidence between consecutive frames. Experimental results show that the four circles is the optimal stirring pattern, achieving the lowest overall mean absolute counting error of 4.384 and the highest overall mean counting confidence of 0.721. Compared with constant-speed stirring, adaptive-speed stirring reduces task execution time by up to 44.7\% and achieves more stable performance across different pest density scenarios. Moreover, the proposed pest counting method reduces the mean absolute counting error by up to 3.428 compared to the single static image counting method under high-density scenarios where occlusion is severe.

\end{abstract}

\section{INTRODUCTION}

Pests can damage crops by feeding and transmitting viruses, leading to yield reduction and significant economic losses. Therefore, timely and accurate monitoring of pest population dynamics and taking appropriate measures are crucial. Water traps, as one of the most widely used pest monitoring tools, are commonly applied to monitor pest population dynamics. Traditionally, monitoring pest populations in water traps requires collecting samples from the field and transferring them to the lab for manual counting under a microscope, which is labor-intensive and time-consuming. While image-based automatic pest counting methods have been explored \cite{chakrabarty2026application}, they commonly rely on single static images, preventing the observation of 
occluded pests and leading to inaccurate counts. To address this challenge, in our previous work \cite{gao2024interactive} \cite{gao2025counting}, we proposed a pest counting method in water traps leveraging interactive stirring and counting confidence assessment, where stirring redistributes pests to reveal occluded individuals for counting, and counting confidence assessment quantifies counting uncertainty to produce more reliable counting results. However, the proposed method depends on manual stirring, where the stirring pattern, speed, and duration are determined by human intuition, introducing subjectivity and variability that can lead to under- or over-stirring, ultimately affecting pest counting accuracy. Although some studies have explored automated stirring \cite{eggl2022mixing, liu2024enhanced, zhang2022enhancement, sochacki2021closed,luo2024intelligent,saito2025learning, szymanska2025robotic}, existing approaches suffer from two major limitations. First, they commonly adopt circular stirring patterns and largely overlook the effects of different stirring patterns on stirring performance. Second, they do not implement adaptive-speed stirring in liquid environments, where the stirring speed is adjusted according to the changing state of the stirred material, as continuous closed-loop control in liquid environments remains challenging due to the inherent complexity of fluid dynamics.

To this end, this paper proposes an automated pest counting 
method in water traps through active robotic stirring for 
occlusion handling, with pattern optimization and adaptive 
speed control. Specifically, 1) An automated robotic arm-based stirring system is 
developed for pest counting in water traps. 2) Six stirring 
patterns are designed and evaluated to investigate the effects of different stirring patterns on pest counting performance, and the optimal pattern is selected based on counting error and counting confidence. 3) A heuristic counting confidence-driven 
closed-loop control system is proposed for adaptive-speed 
robotic stirring, which adjusts stirring speed based on 
the average change rate of counting confidence between 
consecutive frames to improve stirring efficiency and 
counting reliability. Our approach can be broadly applied to other object counting tasks in dynamic liquid environments involving robotic manipulation.

\section{Related work}

\subsection{Image-based Automatic Pest Counting}
Image-based automatic pest counting has evolved from early shallow feature–based methods to the currently dominant deep learning–based paradigm. Shallow feature–based methods rely on traditional image processing and hand-crafted feature representations, making them highly sensitive to environmental variations and limiting their real-world applicability. In contrast, deep learning–based methods, including detection-, segmentation-, and density map estimation-based approaches, automatically learn discriminative features from large-scale data, enabling deployment in complex real-world pest counting scenarios. Despite remarkable progress, accurately counting pests under occlusion remains a fundamental and largely unresolved challenge. Detection-based 
approaches rely on bounding box predictions prone to overlap, 
causing adjacent individuals to merge into a single detection 
and leading to undercounting \cite{hansen2022towards}. Segmentation- and density map estimation-based approaches 
mitigate this by avoiding discrete bounding boxes, offering 
greater robustness in moderately dense scenes, but under 
severe occlusion, both struggle to distinguish individual 
pests from overlapping clusters, causing undercounting \cite{bereciartua2022insect} \cite{kargar2024net}.

To overcome undercounting caused by occlusion, our previous work \cite{gao2024interactive} introduced interactive stirring in water traps to reveal occluded pests for counting, motivated by advances in interactive perception for occlusion handling. However, unlike most interactive perception studies that operate in static environments, pest counting in water traps involves an inherently dynamic setting where fluid motion causes pests to move unpredictably, with their poses, positions, and spatial relationships continuously changing over time, resulting in unstable counting performance. Therefore, our subsequent work \cite{gao2025counting} proposed a counting confidence assessment method to assess the reliability of pest counting results under such dynamic conditions. Although these two works together have significantly improved pest counting accuracy under occlusion, the approach relies on manual stirring, where controlling the stirring process based on human visual perception introduces subjective errors, resulting in under- or over-stirring that adversely affects counting accuracy.

\subsection{Counting Uncertainty}
\label{subsec:counting_uncertainty}
Uncertainty estimation has become a research focus in many critical fields. Current research on uncertainty estimation has made significant progress in probabilistic object detection \cite{hall2020probabilistic}. However, unlike probabilistic object detection, which estimates uncertainty for each detected instance, counting uncertainty in counting tasks should be evaluated from a scene-level perspective, considering not only detected instances but also those instances that were missed detections, as well as the spatial distribution and relationships among all instances, such as density, degree of overlap, and clustering patterns that fundamentally affect countability. Although a few recent studies \cite{oh2020crowd} \cite{xu2024uncertainty} attempt to extend uncertainty modeling from probabilistic object detection to density map estimation-based counting that leverage the fact that density map estimation networks inherently process all pixels in an image, they overlook the spatial distribution and relationships of instances, as density map estimation-based networks lack explicit instance-level localization. To overcome these limitations, our previous work \cite{gao2025counting} proposed a counting confidence assessment method that jointly considers multiple factors affecting counting uncertainty from a scene-level perspective, which is adopted in this paper to assess counting uncertainty of pest counting results and guide the design of adaptive-speed robotic stirring, with further details presented in Section~\ref{sec:materials and methods}.

\subsection{Automated Stirring}
\label{subsec:automated stirring}

To design effective robotic stirring strategies for pest counting, it is necessary to examine existing research on automated stirring, which has primarily focused on two domains: automated stirring for liquid–liquid or solid–liquid mixing, and automated stirring in robotic cooking.

Automated stirring for liquid–liquid or solid-liquid mixing has been widely applied in chemical engineering, pharmaceutical manufacturing, and food processing. Accordingly, a series of studies have focused on optimizing stirring strategies to enhance mixing homogeneity and reduce mixing time, primarily from three aspects: stirrer geometry, stirring speed, and stirring direction. Regarding stirrer geometry, non-traditional and irregularly shaped stirrers can significantly enhance mixing performance by generating complex vortices \cite{eggl2022mixing}. Regarding stirring speed, applying precomputed non-periodic chaotic stirring speed sequences has been shown to substantially reduce mixing time compared to constant-speed stirring \cite{liu2024enhanced}. Regarding stirring direction, bidirectional stirring has been shown to achieve shorter mixing times compared to unidirectional stirring \cite{zhang2022enhancement}. However, these studies have several limitations. First, they largely overlook the effects of different stirring patterns, as conventional fixed-axis stirrers are mechanically confined to circular patterns. Second, although precomputed variable-speed strategies outperform constant-speed stirring, they operate in an open-loop manner rather than in a closed-loop control mechanism that adjusts stirring speed based on the changing state of the stirred material, preventing the achievement of optimal mixing efficiency. This limitation arises because the inherent fluid dynamics of liquids pose significant challenges for the design and implementation of effective closed-loop control. Additionally, most studies simplify the problem by treating the mixing target as a whole and conducting experiments in simulated fluid environments rather than real liquid environments.

In contrast, automated stirring in robotic cooking is commonly designed as a closed-loop control framework, which is feasible because cooking ingredients in robotic cooking are mostly solids or semi-fluids whose states change relatively slowly and can be effectively monitored by contact-based sensors. For instance, Sochacki et al.~\cite{sochacki2021closed} proposed a closed-loop robotic cooking system that uses salinity sensor feedback to control salt addition and stirring during scrambled egg cooking, with each stirring action following a predetermined pattern for a fixed duration. Luo et al.~\cite{luo2024intelligent} proposed a pancake-making robot that performs a fixed-duration preliminary stirring phase with four stirring motions (quick stirring, fine stirring, edge scraping, and whisk shaking) to mix raw ingredients, followed by a perceptive stirring phase that uses force–torque feedback to determine batter uniformity and adaptively stop stirring. Saito et al.~\cite{saito2025learning} designed a robotic cooking system that dynamically reweighted visual, tactile, and torque sensor inputs to monitor egg state and adapt stirring motions accordingly, starting with circular stirring, followed by large-area mixing, flipping, and cutting. However, very few studies have explored automated stirring in robotic cooking within liquid environments. Szymańska and Hughes~\cite{szymanska2025robotic} proposed a robotic coffee-making system that adjusted stirring duration based on clump sizes detected via image processing, but the closed loop was executed only once rather than continuously, highlighting the challenge of continuous closed-loop control in dynamic liquid environments. Overall, existing research on automated stirring in robotic cooking neither investigates the effects of different stirring patterns nor implements adaptive-speed stirring, and treats all ingredients as a unified whole, making state monitoring relatively straightforward.

\section{Materials and methods}
\label{sec:materials and methods}

\subsection{System Overview}

A system overview of the proposed pest counting method is shown in Fig. \ref{fig:system overview}.

\begin{figure}[h]
  \centering
  \includegraphics[width=\linewidth]{figures/overview_of_system.jpg}
  \caption{System overview of the proposed pest counting method.}
  \label{fig:system overview}
\end{figure}

As shown in Fig. \ref{fig:system overview}, the system consists of four main components: 1) a robotic stirring system, 2) pest detection and counting, 3) pest counting confidence assessment, and 4) final pest count computation. The robotic arm performs stirring motions to redistribute pests in the yellow water trap, while the camera simultaneously captures images that are fed into a pest detection network for pest detection and counting, and into a counting confidence assessment model for counting uncertainty estimation. The stirring speed is adaptively adjusted based on the average change rate of counting confidence across the 
most recent $k$ consecutive frames. This process repeats 
iteratively until the average change rate of counting confidence across the most recent $k$ consecutive frames falls below a predefined threshold, at which point 
stirring is terminated, yielding a complete image sequence $\mathrm{Img} = \{\mathrm{Img}_1, \mathrm{Img}_2, \ldots, \mathrm{Img}_n\}$ along with a corresponding counting sequence $N = \{N_1, N_2, \ldots, N_n\}$ and a counting confidence sequence $C = \{C_1, C_2, \ldots, C_n\}$. Finally, the final pest count is computed as a weighted sum of counting results from the complete counting sequence, where each weight $w_i$ is derived from the corresponding counting confidence score.

\subsection{Robotic Stirring System}

The robotic stirring system is composed of four parts: a yellow water trap containing bionic pests, a camera, a robotic arm with a stirring stick, and a PC. The camera is placed above the yellow water trap to capture images of the trap area from a top-down view. The robotic arm is equipped with a wooden stirring stick attached to its end effector, which performs stirring motions within the water trap to redistribute pests. The PC serves as the main controller, coordinating image acquisition and robot motion. In our implementation, a Franka robotic arm is used, controlled via the Franka arm controller connected to the PC through Ethernet, running ROS 2 Humble with MoveIt 2 for motion planning and control.

\subsection{Pest Detection and Counting}
\label{subsec:Pest Detection and Counting}

Pest detection and counting are performed using an improved 
YOLOv5 architecture with three key modifications targeting small pest detection. First, ODConv \cite{li2022omni} is integrated into the backbone to dynamically adjust convolutional kernels across four dimensions, enhancing feature extraction in complex and dynamic scenes. Second, the CoT3 block \cite{li2022contextual} is inserted 
into the backbone to combine local CNN feature extraction 
with Transformer-based global contextual reasoning, 
improving the model's overall learning capability. Third, Soft-NMS is employed in place of standard NMS as the post-processing strategy, improving the handling of densely distributed pests. The model was trained and evaluated on a pest dataset of 410 images, achieving an AP@0.5 of 97.1\% on the test set. Further details are provided in \cite{gao2025counting}.

\subsection{Pest Counting Confidence Assessment}
\label{subsec:Pest Counting with Confidence Assessment}

Pest counting confidence is assessed by a polynomial 
regression model, where the independent variables consist 
of several key factors extracted from the pest image 
that affect counting uncertainty, including counting 
result-related information (the mean detection confidence of all predicted bounding boxes and the predicted number of pests, both obtained from the pest detection model), image quality (assessed via NIQE \cite{mittal2012making}), image complexity (assessed via entropy), image clarity (quantified by average gradient magnitude), and pest distribution uniformity (estimated using a DBSCAN-based clustering algorithm). The dependent variable is the counting confidence of the corresponding pest image, quantified using the Jaccard index ($TP/(TP+FP+FN)$) by comparing detection predictions 
with ground truth labels. The model was trained and evaluated on 890 pest images, achieving an MSE of 0.0028 and $R^2$ of 0.7765. Further details are provided in 
\cite{gao2025counting}.

\begin{figure*}[t]
  \centering
  \includegraphics[width=\textwidth]{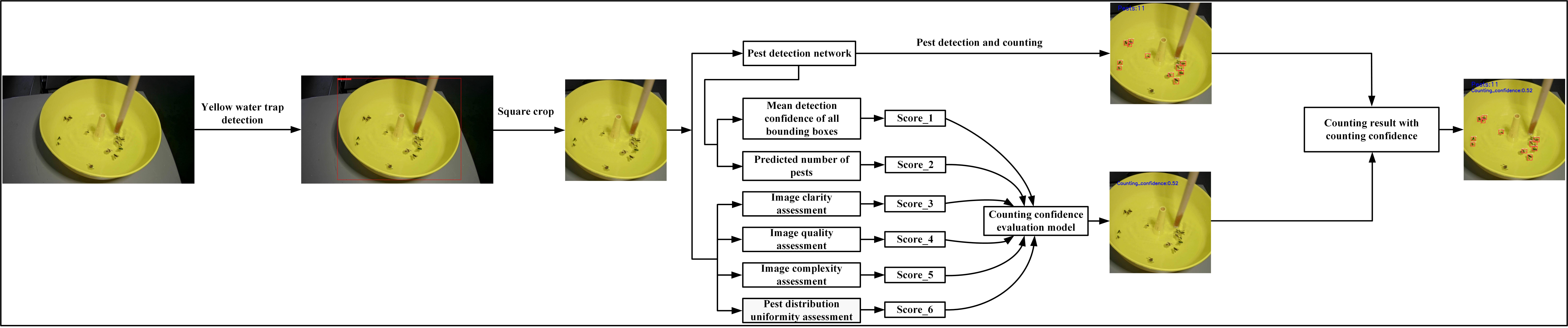}
  \caption{Pipeline of pest counting and counting confidence assessment.}
  \label{fig:counting confidence evaluation}
\end{figure*}

To apply the pest detection and counting method described in Subsection~\ref{subsec:Pest Detection and Counting} and the counting confidence assessment method described above to the proposed robotic stirring-based pest counting, the workflow is as follows: GroundingDINO \cite{liu2024grounding} is first employed to detect and crop the yellow water trap region from the captured image using the text prompt ``yellow pan'', which is then passed to the pest detection model for pest counting. It is worth noting that since the YOLOv5-based detection model requires a square input, the detected yellow water trap region is converted to a square crop centered at the bounding box center with the shorter side as the side length before being passed to the pest detection model. The aforementioned key factors influencing counting uncertainty are then extracted and input into the pre-trained counting confidence assessment model to predict the counting confidence score, as defined in Eq.~\eqref{eq:counting_confidence_prediction}, yielding a pest counting result with an associated counting confidence score, as illustrated in Fig. \ref{fig:counting confidence evaluation}. 

\begin{equation}
\begin{aligned}
C_{\text{predict}} 
&= \beta_0 
+ \sum_{i=1}^{6} \beta_i x_i \\
&\quad + \sum_{i=1}^{6} \beta_{ii} x_i^2 \\
&\quad + \sum_{1 \le i < j \le 6} \beta_{ij} x_i x_j \\
&\quad + \varepsilon
\end{aligned}
\label{eq:counting_confidence_prediction}
\end{equation}

In Eq.~\eqref{eq:counting_confidence_prediction}, each $x_i$ represents a factor influencing the counting confidence. The coefficients $\beta_0$, $\beta_i$, $\beta_{ii}$, $\beta_{ij}$ and $\varepsilon$ are determined through the polynomial 
regression analysis.

\subsection{Final Pest Count Computation}
\label{subsec:Final Pest Count Computation}

By repeating the pest detection and counting, and counting confidence assessment at each time instant throughout a complete stirring sequence, a complete image sequence $\mathrm{Img} = \{\mathrm{Img}_1, \mathrm{Img}_2, \ldots, \mathrm{Img}_n\}$, a corresponding counting sequence $N = \{N_1, N_2, \ldots, N_n\}$ and a counting confidence sequence $C = \{C_1, C_2, \ldots, C_n\}$ are obtained. The final pest count is then computed as a weighted sum of counting results from all sequential images, as shown in Eq.~\eqref{eq:final_count}, where the counting result of each image $N_i$ is weighted by a softmax probability derived from its corresponding counting confidence $C_i$. This approach effectively reduces the impact of uncertain measurements and improves the reliability of pest counting under dynamic liquid conditions.

\begin{equation}
\mathrm{FC}
=
\sum_{i=1}^{n}
\left(
\frac{e^{C_i}}{\sum_{j=1}^{n} e^{C_j}}
\cdot N_i
\right)
\label{eq:final_count}
\end{equation}

\subsection{Selection of Optimal Stirring Pattern}

\begin{figure}[h]
  \centering
  \includegraphics[width=\linewidth]{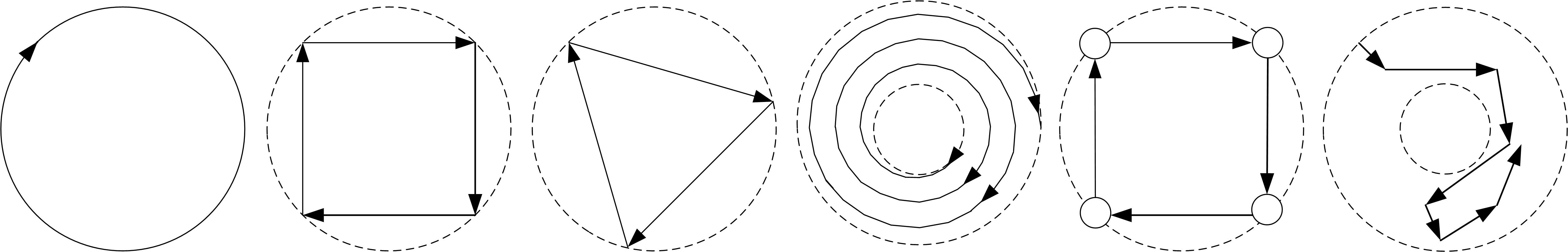}
  \caption{Six designed stirring patterns, shown from left to right: circle, square, triangle, spiral, four circles, and random lines.}
  \label{fig:stirring patterns}
\end{figure}

To investigate the effects of different stirring patterns on pest counting and identify the optimal one, we designed six stirring patterns based on the circular stirring pattern: circle, square, triangle, spiral, four circles, and random lines, as illustrated in Fig. \ref{fig:stirring patterns}. The optimal pattern was selected based on the lowest average counting error and highest average counting confidence (defined in Subsection~\ref{subsec:Evaluation Metrics}). 

As shown in Fig. \ref{fig:stirring patterns}, the circular pattern serves as a reference with a radius of 8 cm, determined by the size of the yellow water trap. The square and triangular patterns are the largest inscribed square and triangle within the reference circle, respectively. The spiral pattern consists of three equally spaced layered spirals starting 3 cm from the center, where the offset accounts for the support column (radius 2 cm) and stirring stick (radius 1 cm) to avoid collisions. The four circles pattern consists of four circles of radius 1 cm evenly distributed along the reference circle, connected by transitional line segments and stirred sequentially. The random lines pattern is formed by multiple randomly generated straight lines within the reference circle, excluding the central region within a 3 cm radius to avoid collisions.

\subsection{Adaptive-Speed Stirring via Closed-Loop Control} 

\begin{figure*}[t]
  \centering
  \includegraphics[width=\textwidth]{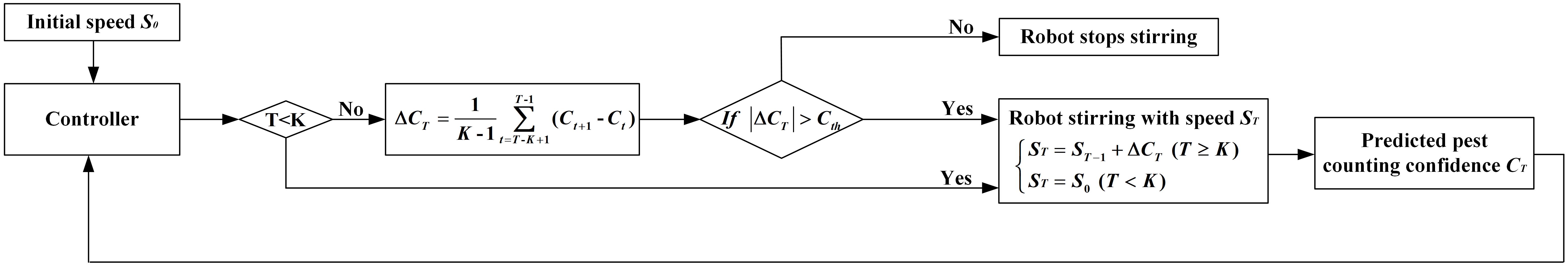}
  \caption{The closed-loop control system for adaptive-speed stirring.}
  \label{fig:closed-loop control}
\end{figure*}

On the basis of selecting the optimal stirring pattern, in order to further improve stirring efficiency, we propose an adaptive-speed stirring strategy implemented through a heuristic counting confidence-driven closed-loop control system, as illustrated in Fig. \ref{fig:closed-loop control}.

As shown in Fig. \ref{fig:closed-loop control},
the robot begins executing the stirring motion along the optimal stirring pattern selected above at an initial stirring speed $S_0$ at time $T_0$. Simultaneously, the camera continuously captures images of the yellow water trap region for pest counting and counting confidence computation until time $T_k$, at which point the average change rate of the pest counting confidence $\Delta C_T$ is calculated. Subsequently, $\Delta C_T$ is compared with a predefined threshold $C_{th}$. If $|\Delta C_T| > C_{th}$, $\Delta C_T$ is used as feedback to adjust the stirring speed according to formula $S_T = S_{T-1} + \Delta C_T$. If $|\Delta C_T| \le C_{th}$, it is considered that the change in pest counting confidence across the most recent $k$ consecutive frames is sufficiently small, and stirring is terminated. This process is repeated iteratively, forming a closed-loop control system in which the variation in pest counting confidence continuously adjusts the stirring speed until the robotic arm stops stirring.

It is worth noting that $\Delta C$ is used here as a heuristic indicator to adjust stirring speed, rather than as a precisely modeled control signal characterizing the explicit relationship between stirring speed, pest visibility, and liquid dynamics. This is because the complex fluid dynamics of the liquid environment in pest counting, involving turbulence, surface waves, and inter-pest interactions, make explicit physical modeling computationally prohibitive. As discussed in Subsection~\ref{subsec:automated stirring}, existing automated stirring studies in liquid environments are predominantly conducted in simulation settings, where fluid state can be directly accessed and controlled. In contrast, using $\Delta C$ derived directly from image observations offers a lightweight and practical alternative that reflects the effect of stirring on pest visibility and liquid dynamics without requiring explicit fluid state estimation. Specifically, a large positive $\Delta C$ indicates that recent stirring has 
successfully revealed occluded pests with limited visual 
disturbance from liquid dynamics, suggesting that increasing stirring speed may remain beneficial. In contrast, a large negative $\Delta C$ implies that excessive liquid motion has degraded counting confidence, warranting a reduction in stirring speed. When $\Delta C$ is very small, further stirring is unlikely to yield additional improvements in pest visibility, and the stirring process is terminated.

\section{Experiments and results}

\subsection{Evaluation Metrics}
\label{subsec:Evaluation Metrics}

Two different sets of metrics are used in this paper. For stirring strategy optimization experiments, which cover the selection of the optimal stirring pattern and adaptive-speed stirring, two metrics are used: the mean absolute counting error ($\text{MAE}_1$) and the mean counting confidence ($\text{MCC}$). For evaluating pest counting performance of the proposed method, the mean absolute counting error ($\text{MAE}_2$) is used. The key distinction is that $\text{MAE}_1$ and $\text{MCC}$ 
are computed by directly comparing ground truth annotations 
with raw detection results on a per-image basis, while 
$\text{MAE}_2$ is computed using the final pest count 
obtained from a complete counting sequence 
(Eq.~\eqref{eq:final_count}), which depends on the 
predicted outputs of both the pest detection model and 
the counting confidence assessment model, as detailed 
below.

\subsubsection{Metrics for Stirring Strategy Optimization}
During robotic stirring, the objective is to redistribute pests to reveal occluded individuals for counting. In addition, it is essential to consider the reliability of counting results, as the inherent fluid dynamics of the liquid, compounded by stirring actions, cause continuous dynamic changes in the pest counting scene, affecting the stability of the counting model. Therefore, two metrics are used to evaluate pest counting 
performance in the stirring strategy optimization 
experiments: the mean absolute error ($\text{MAE}_1$) and the mean counting confidence ($\text{MCC}$), as defined in Eq.~\eqref{eq:mae1} and Eq.~\eqref{eq:mcc}, respectively.

\begin{equation}
\text{MAE}_1 = \frac{1}{M}\sum_{i=1}^{M} |GT_i - TP_i|
\label{eq:mae1}
\end{equation}

\begin{equation}
\text{MCC} = \frac{1}{M}\sum_{i=1}^{M} 
\frac{TP_i}{TP_i + FP_i + FN_i}
\label{eq:mcc}
\end{equation}

In Eq.~\eqref{eq:mae1} and Eq.~\eqref{eq:mcc}, $M$ denotes the total number of images, $GT_i$ represents the actual number of pests in the $i$-th image, and $TP_i$, $FP_i$, $FN_i$ denote the numbers of true positives, false positives, and false negatives produced by the pest detection model for the $i$-th image, respectively. 

Two critical considerations underlying these metrics warrant explanation. First, $\text{MAE}_1$ in Eq.~\eqref{eq:mae1} is defined as the mean of $|GT_i - TP_i|$ across all images, rather than $|GT_i - N_{\text{predicted},i}|$, where 
$N_{\text{predicted},i} = TP_i + FP_i$. This is because the latter does not explicitly distinguish between different types of detection errors: false positives can 
numerically compensate for missed detections, leading to an underestimation of the true counting error. Second, $\text{MCC}$ in Eq.~\eqref{eq:mcc} is computed as the mean of the Jaccard index values across all images, where the Jaccard index serves as the per-image counting confidence, consistent with the definition used to develop the counting confidence assessment model, rather than using the predicted confidence scores from the pre-trained model (Eq.~\eqref{eq:counting_confidence_prediction}). Together, these two design choices ensure that $\text{MAE}_1$ 
and $\text{MCC}$ defined here are intentionally defined to exclude both the predicted pest counts produced by the detection model and the predicted counting confidence generated by the counting confidence assessment model. This is because the primary goal of stirring strategy optimization is to compare the effect of different stirring strategies on counting performance. Since neither the detection model nor the counting confidence assessment model can achieve perfect prediction accuracy, directly incorporating their predicted outputs into the evaluation metrics would introduce additional model-induced errors that conflate with the effect of stirring strategies, potentially masking true differences between different stirring strategies and leading to suboptimal conclusions. 

\subsubsection{Metric for Pest Counting}
For evaluating the pest counting performance of the 
proposed pest counting method, the mean absolute error ($\text{MAE}_2$) is used, as defined in Eq.~\eqref{eq:mae2}, where $S$ denotes 
the total number of counting sequences, $GT_i$ 
represents the actual pest count for the $i$-th counting 
sequence, and $FC_i$ denotes the final pest counting result for the $i$-th counting sequence, computed using Eq.~\eqref{eq:final_count}. Unlike $\text{MAE}_1$, $\text{MAE}_2$ in Eq.~\eqref{eq:mae2} inherently depends on the predicted outputs of both the pest detection model and the counting confidence assessment model, as $FC_i$ is computed as a weighted sum of predicted counting results from all images in the $i$-th counting sequence, where each weight is determined by the corresponding predicted counting 
confidence. 

\begin{equation}
\text{MAE}_2 = \frac{1}{S}\sum_{i=1}^{S} |GT_i - FC_i|
\label{eq:mae2}
\end{equation}

\subsection{Experimental Implementation}

\subsubsection{Selection of Optimal Stirring Pattern}
\begin{figure}[h]
  \centering
  \includegraphics[width=\linewidth]{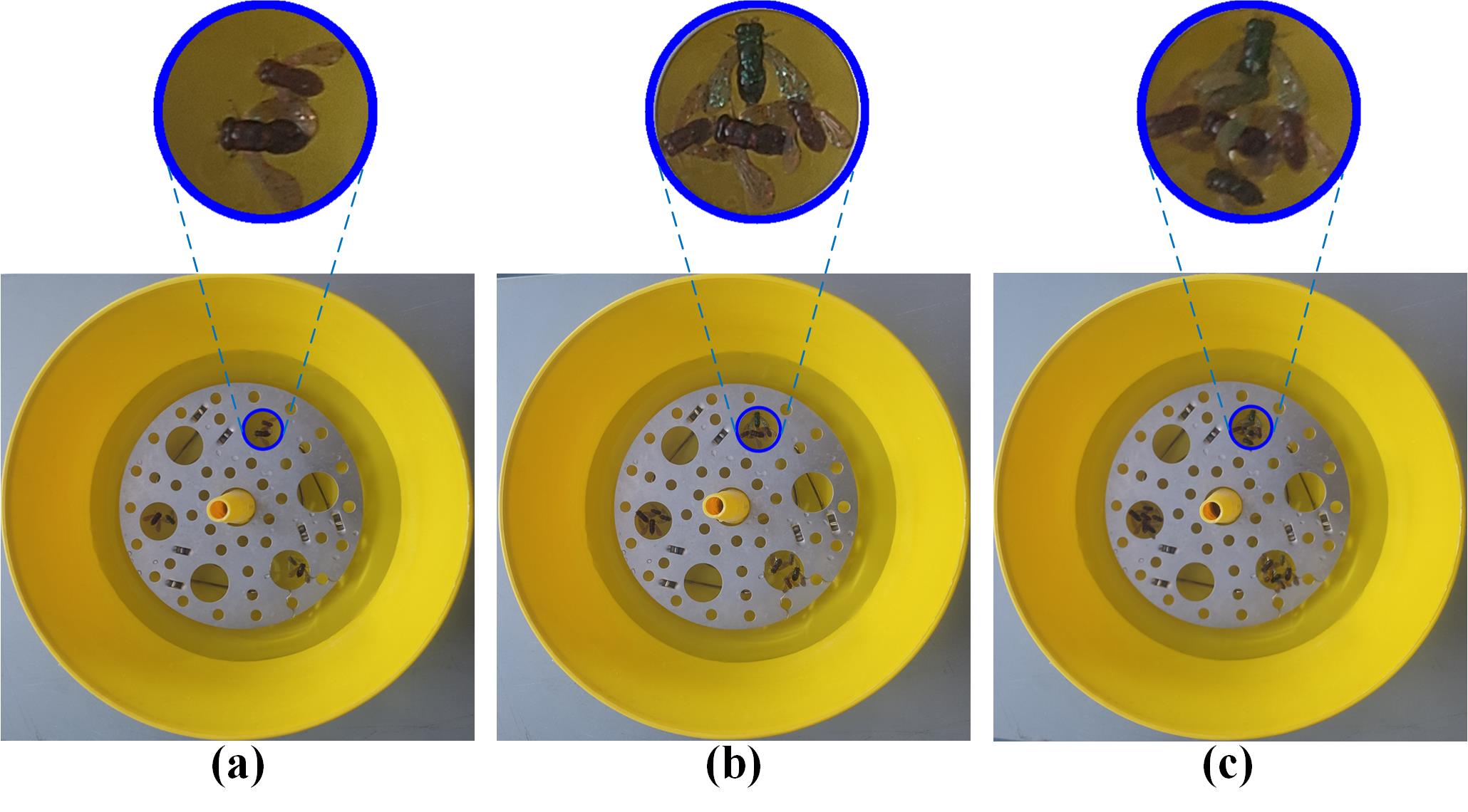}
  \caption{Initial pest arrangements in yellow water traps under three density scenarios: (a) low density, (b) medium density, and (c) high density.}
  \label{fig:setup}
\end{figure}

The complete data collection process is as follows. Starting at 0s, the camera captures the first frame. At 1s, the stirring stick begins to descend, reaching into the yellow water trap by 2s and initiating stirring. Stirring continues until 15s, after which the stirring stick is withdrawn. The camera continues capturing images until 50s, when the water surface becomes nearly calm, with one frame captured every 2s throughout the entire process. All six stirring patterns were sequentially applied to stir pests, with each repeated 20 times for data collection. To ensure fairness, a stainless steel template was used before each trial to fix the initial pest arrangement, with the water surface kept flush with the template, ensuring consistent initial pest distribution across all trials. Three pest density scenarios were established: low, medium, and high, as shown in Fig. \ref{fig:setup}. In the low-density scenario, each local region contained 2 pests with negligible occlusion; in the medium-density scenario, 4 pests with moderate occlusion; and in the high-density scenario, 6 pests with severe occlusion. For each density scenario, all six patterns were sequentially applied, each repeated 20 times for data collection.

\subsubsection{Adaptive-Speed Stirring}
To evaluate the proposed adaptive-speed robotic stirring strategy, two comparative groups were designed. In the first group, the robotic arm performed stirring at a constant speed $S = 0.5$, where 0.5 is a scaling factor relative to the maximum movement speed of the Franka arm. In the second group, the robotic arm employed the proposed closed-loop control to adaptively adjust the stirring speed, with an initial speed $S_0 = 0.5$ and a threshold $C_{th} = 0.01$. $K$ is set to 3, meaning it always calculates the average change rate of the counting confidence of the captured images over the most recent three time instants. Both the threshold $C_{th}$ and $K$ were empirically determined. It is worth noting that in the constant-speed group, the system still continuously computes $\Delta C$ and terminates stirring when $|\Delta C| \le C_{th}$, consistent with the adaptive-speed group. Under each of the three pest density scenarios (low, medium, and high), both stirring strategies were alternately executed for 20 trials each, using the same stainless steel template to ensure consistent initial pest distributions.

\subsection{Selection of Optimal Stirring Pattern}

Under each of the three pest density scenarios, all six stirring patterns were applied for 20 trials each. For each trial, the $\text{MAE}_1$ and $\text{MCC}$ were computed. These were then averaged across 20 trials per pattern per density scenario, and further aggregated across all three density scenarios to obtain the overall average $\text{MAE}_1$ and $\text{MCC}$ for each pattern. The results are summarized in Table \ref{tab:pattern_comparison}.

\begin{table}[h]
\caption{Comparison of $\text{MAE}_1$ and $\text{MCC}$ under different stirring patterns.}
\label{tab:pattern_comparison}
\centering
\begin{tabular}{|
>{\centering\arraybackslash}p{2cm}|
>{\centering\arraybackslash}p{2.1cm}|
>{\centering\arraybackslash}p{1.4cm}|
>{\centering\arraybackslash}p{1.5cm}|
}
\hline
\textbf{Density} & \textbf{Pattern} & $\textbf{MAE}_1$ & \textbf{MCC} \\
\hline

\multirow{6}{*}{Low}
& circle             & 1.027 & 0.870 \\ \cline{2-4}
& square             & 1.160 & 0.879 \\ \cline{2-4}
& triangle           & 1.040 & 0.853 \\ \cline{2-4}
& spiral             & 1.050 & 0.847 \\ \cline{2-4}
& \textbf{four circles} & {\bfseries 0.964} & {\bfseries 0.880} \\ \cline{2-4} 
& random lines       & 1.000 & 0.879 \\ \hline

\multirow{6}{*}{Medium}
& circle             & 4.064 & 0.700 \\ \cline{2-4}
& square             & 3.596 & 0.727 \\ \cline{2-4}
& triangle           & 3.921 & 0.699 \\ \cline{2-4}
& spiral             & 3.589 & 0.727 \\ \cline{2-4}
& four circles & 3.633 & 0.726 \\ \cline{2-4}
& \textbf{random lines}       & \textbf{3.439} & \textbf{0.750} \\ \hline

\multirow{6}{*}{High}
& circle             & 9.829 & 0.479 \\ \cline{2-4}
& square             & 8.889 & 0.536 \\ \cline{2-4}
& triangle           & 9.021 & 0.519 \\ \cline{2-4}
& spiral             & 9.558 & 0.486 \\ \cline{2-4}
& \textbf{four circles} & \textbf{8.556} & \textbf{0.556} \\ \cline{2-4}
& random lines       & 9.404 & 0.517 \\ \hline

\multirow{6}{*}{Overall average}
& circle             & 4.973 & 0.683 \\ \cline{2-4}
& square             & 4.548 & 0.714 \\ \cline{2-4}
& triangle           & 4.661 & 0.690 \\ \cline{2-4}
& spiral             & 4.732 & 0.687 \\ \cline{2-4}
& \textbf{four circles} & \textbf{4.384} & \textbf{0.721} \\ \cline{2-4}
& random lines       & 4.614 & 0.715 \\ \hline

\end{tabular}
\end{table}

As shown in Table \ref{tab:pattern_comparison}, four circles achieves the lowest $\text{MAE}_1$ and highest $\text{MCC}$ at both low density ($\text{MAE}_1$: 0.964, $\text{MCC}$: 0.880) and high density ($\text{MAE}_1$: 8.556, $\text{MCC}$: 0.556), while random lines performs best at medium density ($\text{MAE}_1$: 3.439, $\text{MCC}$: 0.750). Since pest density varies randomly in practice, the optimal pattern is determined based on overall average performance across all three density levels. As shown in the last row of Table \ref{tab:pattern_comparison}, the four circles achieves the lowest overall average $\text{MAE}_1$ (4.384) and the highest overall average $\text{MCC}$ (0.721), and is therefore selected as the optimal stirring pattern. Notably, the commonly used circular pattern performed the worst, with the highest overall average $\text{MAE}_1$ (4.973) and the lowest overall average $\text{MCC}$ (0.683).

\subsection{Adaptive-Speed Stirring}
\label{subsec:Evaluation of Adaptive-Speed Stirring}

Using the optimal four circles pattern, 20 trials of both constant-speed and adaptive-speed stirring were conducted under each of the three pest density scenarios. For each trial, the task execution time (from stirring start to termination), $\text{MAE}_1$, and $\text{MCC}$ were recorded. The average and standard deviation of task execution time, along with the $\text{MAE}_1$ and $\text{MCC}$ across 20 trials, were then computed for both stirring strategies under each density scenario. The task execution time results are summarized in Table \ref{tab:time_consumed}. The $\text{MAE}_1$ and $\text{MCC}$ results are presented in Table \ref{tab:speed_comparison}.

\begin{table}[h]
\caption{Comparison of task execution time under constant and adaptive stirring speeds.}
\label{tab:time_consumed}
\centering
\begin{tabular}{|
>{\centering\arraybackslash}p{1.4cm}|
>{\centering\arraybackslash}p{1.1cm}|
S[table-format=2.2(2.2)]|
}
\hline
\textbf{Density} & \textbf{Speed} & {\textbf{Task execution time [s] (mean $\pm$ std)}} \\
\hline

\multirow{2}{*}{Low} 
& constant & 16.6 \pm 10.2 \\ \cline{2-3}
& adaptive & 10.1 \pm 4.8 \\ \hline

\multirow{2}{*}{Medium} 
& constant & 20.8 \pm 15.8 \\ \cline{2-3}
& adaptive & 11.5 \pm 3.5 \\ \hline

\multirow{2}{*}{High} 
& constant & 18.9 \pm 11.3 \\ \cline{2-3}
& adaptive & 12.0 \pm 3.4 \\ \hline

\end{tabular}
\end{table}

\begin{table}[h]
\caption{Comparison of $\text{MAE}_1$ and $\text{MCC}$ under constant and adaptive stirring speeds.}
\label{tab:speed_comparison}
\centering
\begin{tabular}{|
>{\centering\arraybackslash}p{1.4cm}|
>{\centering\arraybackslash}p{1.1cm}|
>{\centering\arraybackslash}p{2cm}|
>{\centering\arraybackslash}p{2cm}|
}
\hline
\textbf{Density} & \textbf{Speed} & $\textbf{MAE}_1$ & \textbf{MCC} \\
\hline

\multirow{2}{*}{Low} 
& constant & 2.313 & 0.681 \\ \cline{2-4}
& adaptive & 2.314 & 0.713 \\ \hline

\multirow{2}{*}{Medium} 
& constant & 5.972 & 0.547 \\ \cline{2-4}
& adaptive & 6.006 & 0.592 \\ \hline

\multirow{2}{*}{High} 
& constant & 11.378 & 0.399 \\ \cline{2-4}
& adaptive & 11.947 & 0.423 \\ \hline

\end{tabular}
\end{table}

As shown in Table \ref{tab:time_consumed}, adaptive-speed stirring demonstrates significant advantages in efficiency and stability compared to constant-speed stirring across all density scenarios. The average task execution time is reduced by 39.2\%, 44.7\%, and 36.5\% under low, medium, and high density, respectively, while the standard deviation is reduced by 52.9\%, 77.8\%, and 69.9\%.

As shown in Table \ref{tab:speed_comparison}, the adaptive-speed stirring achieves comparable $\text{MAE}_1$ to constant-speed stirring across all three density scenarios, with slight differences of 0.001, 0.034, and 0.569 for low, medium, and high density, respectively. This finding validates the rationality of using counting confidence change rate as the stirring termination criterion: since both strategies terminate stirring when $|\Delta C| \le C_{th}$, they converge to similar pest redistribution states and thus yield comparable counting errors. Notably, the adaptive-speed stirring achieves consistently higher $\text{MCC}$, with improvements of 4.7\%, 8.2\%, and 6.0\% for low, medium, and high density scenarios, respectively. This is attributed to the closed-loop control mechanism: when $\Delta C$ is positive, the stirring speed automatically increases to promote pest redistribution, whereas when $\Delta C$ is negative, the stirring slows down to reduce unnecessary disturbance. In contrast, constant-speed stirring continuously disturbs the water regardless of pest distribution, adversely affecting image quality and counting reliability.

\subsection{Pest Counting under Robotic Stirring}

To evaluate the effectiveness of pest counting under robotic stirring, we compared the $\text{MAE}_2$ across 20 trials under different pest density scenarios, obtained using three methods: the single static image counting method (Static), the proposed counting method under constant-speed robotic stirring (Constant), and the proposed counting method under adaptive-speed robotic stirring (Adaptive), based on the experiments conducted in Subsection~\ref{subsec:Evaluation of Adaptive-Speed Stirring}. The single static image counting method employs the same pest detection network as the proposed method, but performs counting only once on an image captured prior to stirring, with the result treated as the final pest counting result for each trial. The proposed method computes the pest count for each complete counting sequence using Eq.~\eqref{eq:final_count}, with the result treated as the final pest counting result for the corresponding trial. The comparison is summarized in 
Table~\ref{tab:counting_comparison}.

\begin{table}[h]
\caption{Comparison of $\text{MAE}_2$ obtained using different counting methods.}
\label{tab:counting_comparison}
\centering
\begin{tabular}{|
>{\centering\arraybackslash}p{1.9cm}|
>{\centering\arraybackslash}p{1.1cm}|
>{\centering\arraybackslash}p{1.1cm}|
>{\centering\arraybackslash}p{1.1cm}|
}
\hline
\textbf{Density} & \textbf{Static} & \textbf{Constant} & \textbf{Adaptive} \\
\hline

Low & 3.050 & 2.105 & 2.186 \\ \hline
Medium & 9.000 & 5.934 & 5.975 \\ \hline
High & 14.500 & 11.072 & 11.452 \\ \hline

\end{tabular}
\end{table}

As shown in Table~\ref{tab:counting_comparison}, several 
conclusions can be drawn. First, compared to the single 
static image counting method, both proposed methods achieve 
substantially lower $\text{MAE}_2$ across all density 
scenarios. Specifically, under constant-speed robotic stirring, $\text{MAE}_2$ is reduced by 0.945, 3.066, and 3.428 for low, medium, and high density, respectively; under adaptive-speed stirring, the reductions are 0.864, 3.025, and 3.048. The improvement is modest at low density, as occlusion is rare and leaves little room for stirring-based improvement, but substantially larger at medium and high density, where severe inter-pest occlusion causes significant undercounting in the static image counting method. Second, $\text{MAE}_2$ is comparable between constant- and adaptive-speed stirring across all densities, consistent with the finding in 
Table~\ref{tab:speed_comparison} that $\text{MAE}_1$ is 
nearly equivalent between the two stirring speed strategies. 
Third, although $\text{MAE}_2$ is significantly reduced 
under robotic stirring compared to the single static image counting method, it remains considerable under high-density scenarios, primarily attributed to three factors: the 
inherent limitations of the detection model, which cannot 
achieve perfect detection accuracy; some physically adhered 
pests cannot be separated even under robotic stirring; and 
temporarily separated pests may re-encounter others due to 
fluid motion, forming new occlusions. In summary, 
simultaneously resolving all occlusions in a dynamically 
changing liquid environment is practically infeasible, 
highlighting the inherent challenge of accurate pest 
counting under such conditions.

\section{CONCLUSIONS}

In this paper, we proposed an automated pest counting 
method in water traps through active robotic stirring 
for occlusion handling. Six stirring patterns were designed and evaluated, with four circles identified as optimal, achieving the lowest overall average $\text{MAE}_1$ (4.384) and highest overall average $\text{MCC}$ (0.721) across different density scenarios, while the commonly used circular pattern performed the worst. A heuristic counting confidence-driven closed-loop control system was further proposed for adaptive-speed stirring, which consistently outperformed constant-speed stirring in time efficiency and stability across different pest density scenarios. Compared to constant-speed stirring, adaptive-speed stirring reduced average task execution time by 39.2\%, 44.7\%, and 36.5\% and decreased its standard deviation by 52.9\%, 77.8\%, and 69.9\% under low, medium, and high pest densities, respectively. Furthermore, compared to the single static image counting 
method, the proposed pest counting method under 
constant-speed robotic stirring reduces $\text{MAE}_2$ 
by 0.945, 3.066, and 3.428 under low, medium, and high 
density, respectively, and under adaptive-speed robotic 
stirring by 0.864, 3.025, and 3.048, respectively, 
demonstrating substantially improved counting accuracy 
under occlusion. However, there are some limitations. First, only six stirring patterns were considered, and more effective ones may exist beyond those evaluated. Nevertheless, this paper establishes a systematic evaluation framework for comparing the effects of different stirring patterns on counting tasks in liquid environments, and future work will explore a wider range of robotic stirring patterns, such as figure-of-eight patterns, star-shaped patterns. Second, the computational cost of counting confidence estimation is relatively high, with processing times of 1–2 seconds per time instant, making true real-time closed-loop control difficult. This latency may cause feedback signals to lag behind rapid pest movements, reducing control responsiveness. Future work will focus on optimizing the counting confidence assessment pipeline to enable more efficient and responsive closed-loop control.

\addtolength{\textheight}{-12cm}   




\bibliography{references}  

@article{chakrabarty2026application,
  title={Application of artificial intelligence in insect pest identification-A review},
  author={Chakrabarty, Sourav and Deb, Chandan Kumar and Marwaha, Sudeep and Haque, Md Ashraful and Kamil, Deeba and Bheemanahalli, Raju and Shashank, Pathour Rajendra},
  journal={Artificial Intelligence in Agriculture},
  volume={16},
  number={1},
  pages={44--61},
  year={2026},
  publisher={Elsevier}
}

@article{hansen2022towards,
  title={Towards machine vision for insect welfare monitoring and behavioural insights},
  author={Hansen, Mark F and Oparaeke, Alphonsus and Gallagher, Ryan and Karimi, Amir and Tariq, Fahim and Smith, Melvyn L},
  journal={Frontiers in Veterinary Science},
  volume={9},
  pages={835529},
  year={2022},
  publisher={Frontiers Media SA}
}

@article{kargar2024net,
  title={Y-net: Insect counting and segmentation using deep learning on embedded devices},
  author={Kargar, Amin and Zorbas, Dimitrios and Gaffney, Michael and O'Flynn, Brendan and Tedesco, Salvatore},
  journal={IEEE Access},
  year={2024},
  publisher={IEEE}
}

@article{bereciartua2022insect,
  title={Insect counting through deep learning-based density maps estimation},
  author={Bereciartua-P{\'e}rez, Arantza and G{\'o}mez, Laura and Pic{\'o}n, Artzai and Navarra-Mestre, Ram{\'o}n and Klukas, Christian and Eggers, Till},
  journal={Computers and Electronics in Agriculture},
  volume={197},
  pages={106933},
  year={2022},
  publisher={Elsevier}
}

@article{gao2024interactive,
  title={Interactive Image-Based Aphid Counting in Yellow Water Traps under Stirring Actions},
  author={Gao, Xumin and Stevens, Mark and Cielniak, Grzegorz},
  journal={arXiv preprint arXiv:2411.10357},
  year={2024}
}

@inproceedings{oh2020crowd,
  title={Crowd counting with decomposed uncertainty},
  author={Oh, Min-hwan and Olsen, Peder and Ramamurthy, Karthikeyan Natesan},
  booktitle={Proceedings of the AAAI conference on artificial intelligence},
  volume={34},
  number={07},
  pages={11799--11806},
  year={2020}
}

@inproceedings{xu2024uncertainty,
  title={Uncertainty-aware Continuous Implicit Neural Representations for Remote Sensing Object Counting},
  author={Xu, Siyuan and Wang, Yucheng and Fan, Mingzhou and Yoon, Byung-Jun and Qian, Xiaoning},
  booktitle={International Conference on Artificial Intelligence and Statistics},
  pages={4105--4113},
  year={2024},
  organization={PMLR}
}

@article{gao2025counting,
  title={Counting with Confidence: Accurate Pest Monitoring in Water Traps},
  author={Gao, Xumin and Stevens, Mark and Cielniak, Grzegorz},
  journal={IFAC-PapersOnLine},
  volume={59},
  number={23},
  pages={233--238},
  year={2025},
  publisher={Elsevier}
}

@article{eggl2022mixing,
  title={Mixing by stirring: Optimizing shapes and strategies},
  author={Eggl, Maximilian F and Schmid, Peter J},
  journal={Physical Review Fluids},
  volume={7},
  number={7},
  pages={073904},
  year={2022},
  publisher={APS}
}

@article{liu2024enhanced,
  title={Enhanced particle mixing performance of liquid-solid reactor under non-periodic chaotic stirring},
  author={Liu, Qiankun and Wang, Shibo and Xu, Jianxin and Sun, Hui and Wang, Hua},
  journal={Chemical Engineering Research and Design},
  volume={211},
  pages={78--94},
  year={2024},
  publisher={Elsevier}
}

@article{zhang2022enhancement,
  title={Enhancement of solid-liquid mixing state quality in a stirred tank by cascade chaotic rotating speed of main shaft},
  author={Zhang, Lian and Yang, Kai and Li, Meng and Xiao, Qingtai and Wang, Hua},
  journal={Powder Technology},
  volume={397},
  pages={117020},
  year={2022},
  publisher={Elsevier}
}

@inproceedings{sochacki2021closed,
  title={Closed-loop robotic cooking of scrambled eggs with a salinity-based ‘taste’sensor},
  author={Sochacki, Grzegorz and Hughes, Josie and Hauser, Simon and Iida, Fumiya},
  booktitle={2021 IEEE/RSJ International Conference on Intelligent Robots and Systems (IROS)},
  pages={594--600},
  year={2021},
  organization={IEEE}
}

@inproceedings{luo2024intelligent,
  title={An intelligent robotic system for perceptive pancake batter stirring and precise pouring},
  author={Luo, Xinyuan and Jin, Shengmiao and Huang, Hung-Jui and Yuan, Wenzhen},
  booktitle={2024 IEEE/RSJ International Conference on Intelligent Robots and Systems (IROS)},
  pages={5970--5977},
  year={2024},
  organization={IEEE}
}

@inproceedings{saito2025learning,
  title={Learning Multimodal Attention for Manipulating Deformable Objects with Changing States},
  author={Saito, Namiko and Tatsumi, Mayu and Kubo, Ayuna and Suzuki, Kanata and Ito, Hiroshi and Sugano, Shigeki and Ogata, Tetsuya},
  booktitle={2025 IEEE-RAS 24th International Conference on Humanoid Robots (Humanoids)},
  pages={460--467},
  year={2025},
  organization={IEEE}
}

@article{szymanska2025robotic,
  title={Robotic optimization of powdered beverages leveraging computer vision and Bayesian optimization},
  author={Szyma{\'n}ska, Emilia and Hughes, Josie},
  journal={Frontiers in Robotics and AI},
  volume={12},
  pages={1603729},
  year={2025}
}

@inproceedings{liu2024grounding,
  title={Grounding dino: Marrying dino with grounded pre-training for open-set object detection},
  author={Liu, Shilong and Zeng, Zhaoyang and Ren, Tianhe and Li, Feng and Zhang, Hao and Yang, Jie and Jiang, Qing and Li, Chunyuan and Yang, Jianwei and Su, Hang and others},
  booktitle={European conference on computer vision},
  pages={38--55},
  year={2024},
  organization={Springer}
}

@article{li2022omni,
  title={Omni-dimensional dynamic convolution},
  author={Li, Chao and Zhou, Aojun and Yao, Anbang},
  journal={arXiv preprint arXiv:2209.07947},
  year={2022}
}

@article{li2022contextual,
  title={Contextual transformer networks for visual recognition},
  author={Li, Yehao and Yao, Ting and Pan, Yingwei and Mei, Tao},
  journal={IEEE transactions on pattern analysis and machine intelligence},
  volume={45},
  number={2},
  pages={1489--1500},
  year={2022},
  publisher={IEEE}
}

@article{mittal2012making,
  title={Making a “completely blind” image quality analyzer},
  author={Mittal, Anish and Soundararajan, Rajiv and Bovik, Alan C},
  journal={IEEE Signal processing letters},
  volume={20},
  number={3},
  pages={209--212},
  year={2012},
  publisher={IEEE}
}

@inproceedings{hall2020probabilistic,
  title={Probabilistic object detection: Definition and evaluation},
  author={Hall, David and Dayoub, Feras and Skinner, John and Zhang, Haoyang and Miller, Dimity and Corke, Peter and Carneiro, Gustavo and Angelova, Anelia and S{\"u}nderhauf, Niko},
  booktitle={Proceedings of the IEEE/CVF Winter Conference on Applications of Computer Vision},
  pages={1031--1040},
  year={2020}
}
\bibliographystyle{IEEEtran}

\end{document}